# A new robust feature selection method using variance-based sensitivity analysis

**Saman Sadeghyan**
College of Engineering, Department of Computer
Engineering, University of Tehran
sadeghyan95@ut.ac.ir

## Abstract

Excluding irrelevant features in a pattern recognition task plays an important role in maintaining a simpler machine learning model and optimizing the computational efficiency. Nowadays with the rise of large scale datasets, feature selection is in great demand as it becomes a central issue when facing high-dimensional datasets. The present study provides a new measure of saliency for features by employing a Sensitivity Analysis (SA) technique called the extended Fourier amplitude sensitivity test, and a well-trained Feedforward Neural Network (FNN) model, which ultimately leads to the selection of a promising optimal feature subset. The paper's ideas are mainly demonstrated based on adopting FNN model for feature selection in classification problems. But in the end, a generalization framework is discussed in order to give insights into the usage in regression problems as well as expressing how other function approximate models can be deployed. Effectiveness of the proposed method is verified by result analysis and data visualization for a series of experiments over several well-known datasets drawn from UCI machine learning repository.

## 1    Introduction

Feature selection is a crucial issue in many classification and regression problems especially when the learning task involves high-dimensional dataset. The ultimate aim is to select the optimal subset from a larger set of features by eliminating features with low or no predictive information. This also may result in the improvement of Machine Learning (ML) model's performance [1].

The approaches exist for feature selection can generally be divided into three groups: filter, wrapper, and embedded approaches [2]. Presented methods with filter approach function regardless of ML model. They basically omit all features which are not able to satisfy a specified criterion. filter approach is very popular particularly because of its computational efficiency. Popular filter methods are F-score criterion [3], mutual information [4], and correlation [5]. On the other hand, the wrapper approach employs a ML model to evaluate each selected feature subset in order to eventually delete irrelevant features. As a result of that, although wrapper approach might provide a better feature subset, but it comparatively acquires way higher CPU time specially when the number of features is large [6]. Embedded approaches that are also known as hybrid approaches, try to exploit both previous discussed approaches in a complementary manner [7]. For further information, feature selection algorithms have been well reviewed in [8, 9].

Sensitivity analysis (SA) is a powerful tool that has been employed in many areas by researchers for investigating the influence of each input factor on output variations in a specified model [10]. However, the term "sensitivity analysis" can be indicated for number of very different problems, such as, parameter screening, global SA or system analysis [11] [12] [13]. In this study, such term particularly means apportioning the model output value uncertainty (e.g., variance), to the uncertainty exists for the input factors, and generally speaking, understanding the activeness of input factors. Results derived from the SA are highly dependent on the input factors, which should be carefully chosen.

In this work, a new method for specifying the saliency of features in ML problems is proposed, where a FNN is considered as the model that its output is studied with respect to input factors which are input features. Unlike most of the feature selection methods that examine whether a feature should be selected in a ML model or not, In the first place, a ranking system is provided for all the existing features which ultimately leads to removal of the identified redundant features in a threshold-based manner. Although, main focus of the present study is on classification problems and multi-output FNN, but it is discussed that how the method can be used for the regression problems as well as expressing how other function approximate models can be employed. Eventually, by analyzing the results derived from a series of experiments over several well-known datasets drawn from (University of California, Irvine) machine learning repository [14], it is confirmed that feature selection based on the proposed measure of saliency for features leads to promising solutions.

The rest of this paper is organized as follows. Section 2 presents an overview of the FNN model and notations. In Section 3, the proposed method for feature selection is discussed in details. In Section 4, the experimental results from the proposed method are analyzed. Finally, Section 5 presents some conclusions and future directions of the paper.

## 2 FNN model and notations

By providing enough hidden units for a FNN model with as few as one hidden layer, using any "squashing" activation function (such as the logistic sigmoid activation function), the network is capable of approximating any measurable function to any desired degree of accuracy [15, 16]. Hence, the network considered in this work without loss of generality is a sigmoidal fully connected feedforward model with one hidden layer. Related calculations and notations are discussed in the following.

Total number of neurons that are incorporated in input and output layers are respectively, H and $K$, which are specified according to the number of features and class labels in given dataset. Also, $N$ as a determinable value by user denotes the number of units in hidden layer. Assume input matrix to be $G = [X_1, X_2, ..., X_L]$, containing $L$ training samples with dimensions of $H \times L$, then $Y_l^{(0)}$ is the output vector of the input layer for the $l_{th}$ sample can be expressed as:

$$Y_l^{(0)} = X_l, \quad (l = 1, ..., L) \tag{1}$$

Since all of the nodes in hidden layer are connected to every node in input layer as well as output layer, $Y_l^{(1)}$ is the output vector of hidden layer that is computed by:

$$Y_l^{(1)} = h(W^T Y_l^{(0)} + B^{(0)}), \quad (l = 1, ..., L) \tag{2}$$

where h is the sigmoid activation function for each hidden unit with the form of $h(x) = 1/(1 + \exp(-x))$, $W$ is the matrix of weights between input and hidden layers with dimensions of $H \times N$, $B^{(0)}$ is the vector of biases that is connected from the bias node to each node in the hidden layer.

In order to network to operate as a classifier model, we want output layer to represent a conditional probability distribution over a discrete variable such as all $K$ target classes in the objective dataset ($C = [c_1, c_2, ..., c_K]$). This is attainable by using the softmax activation function for output layer. Therefore, for $l_{th}$ training sample, while $Y_l^{(2)}$ is output vector of output layer and $P(c_i|X_l)$ represents the probability distribution of the class label $c_i$ (corresponding $i_{th}$ output node), then the calculations of last layer can be expressed as below:

$$Y_l^{(2)} = \text{softmax}(V^T Y_l^{(1)} + B^{(1)}), \quad (l = 1, ..., L) \tag{3}$$

$$P(c_i|X_l) = \frac{\exp(y_i^{(2)})}{\sum_{k=1}^{K} \exp(y_k^{(2)})}, \quad (k = 1, \ldots, K) \tag{4}$$

where $V$ is the matrix of weights between hidden and output layers with dimensions of $N \times K$, $B^{(1)}$ is the vector of biases that is connected from bias node to each node in output layer. $y_i$ and $y_k$ are output values of output unit $i$ and $k$ in $Y_l^{(2)}$. Eventually, using the principle of maximum likelihood determines the predicted class label by the network. For the scheme of training the network in order to learn the output probabilities, we take advantage of Stochastic Gradient Descent (SGD) as one of the most popular gradient-based learning algorithms. It should be noted, for classifier networks with multi-output units, using cross-entropy criterion (i.e., one-hot representation of the label), is proved to be the better approach instead of using mean square error [17, 18]. Altogether, in case of loss function, we try to minimize the Cross-entropy Error (CE) for all training samples:

$$CE = -\frac{1}{L}\sum_{l=1}^{L}\sum_{i=1}^{K} d_i \log P(c_i|X_l) \tag{5}$$

where $d_i$ is the desired output of the $i_{th}$ output unit, for $l_{th}$ training sample. Meanwhile, using the log function here undoes the exponential of the sigmoid functions in output layer, and prevents the saturation problem in gradient-based learning algorithm [19].

## 3     The proposed feature selection method

SA is the study of relative importance of different input factors on the model output [13]. Input factors of the model can be any adjustable quantity in the specification of the model. An input factor can be an initial condition, a parameter, etc. Performing SA of a complex model can disclose contribution of the defined input factors to the variation of model output. Resolving the contribution of each input factor to the global task assigned to the model provides a measure of saliency for them. In the end, the insignificant factors can be eliminated.

Among all of the available methods for SA, sampling method is the most common one that is used in this work [20]. The procedure of implementing a sample-based method basically begins by specifying the model and input factors that are required to be included in the analysis, which is followed by defining probability density functions (ranges of variation), for each input factor. After that, model is evaluated by every sample in a set of samples that are generated accordingly. Eventually, by apportioning the variance of the output according to the input factors, we are able to estimate the contribution of each input factor [21].

In this work however, the main aim is to investigate saliency of each feature using a FNN model. Hence, a normalized contribution percentage for each feature (input unit) is proposed. The feature with higher contribution percentage has higher activeness which is recognized to be more important than features with lower contribution percentage as they have lower influence on network's output.

As mentioned earlier, specification of the model is the first step in the SA that is very influential on the analysis outcome. In case of feedforward network discussed in the previous section network, the overall network output is the resultant of multiple units in output layer. As a result, a model per output unit is defined for the purpose of studying the significance of features. Considering the FNN model with K output units, K defined models are illustrated in Fig 1.

Therefore, a typical model with the mapping function $F(.)$ and with respect to $k_{th}$ output unit is defined as follows:

$$Y_k = F(y_1^{(0)}, y_2^{(0)}, \ldots, y_H^{(0)}) \tag{6}$$

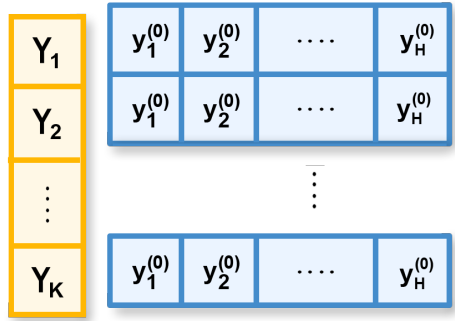

Figure 1: K defined models and their input factors.

In this paper, Extended Fourier Amplitude Sensitivity Test (EFAST), as an efficient sample-based method which has been successfully applied for FNN model [22] is used for generating samples. Also, total Effect (TE) is used which is a comprehensive quantitative measure of SA. TE for $h_{th}$ feature in a typical model like $K$ is computed as bellow:

$$\text{TE}_{hk} = \frac{E_{y_h^{(0)}}[Var(Y|y_h^{(0)})]}{Var(Y)} \tag{7}$$

where $E_{y_h^{(0)}}[Var(Y|y_h^{(0)})]$ is the expected residual variance of model response when all of the other factors vary but $y_h^{(0)}$, and $Var(Y)$ is the variance of model response (more detailed description regarding TE can be found in [20]).

$S_h^T$ that is sum of total effects with respect to all $K$ output units for a feature is taken into account in order to determine the overall contribution in the network. Eventually, the normalized contribution percentage ($C_h$) for the typical feature $h$ is expressed as:

$$C_h = \frac{S_h^T}{\sum_{h=1}^{H} S_h^T} \tag{8}$$

Finally, the steps of the proposed feature selection method can be summarized as follows:

- Initialize the parameters and train the FNN model until the early stopping criteria.
- Define the model and its output and input factors you want to include in your analysis.
- Assign the probability density functions with the ranges of variation to each input factor.
- Generate samples accordingly and evaluate the model.
- Estimate the influences or relative importance of each feature and compute the normalized contribution percentage for all the features.
- Remove input features with low contribution percentage based on a defined threshold.

## 4    Simulations and result analysis

In this section, simulations are conducted using several classification problems with intend to cover examples of small, medium high-dimensional datasets. The proposed method in the paper is implemented in python programming language by exploiting open source libraries in Scipy stack [23]. The final project code is available on [24].

### 4.1 Initialization and parameter setting

For the purpose of early stopping and preventing a resulted network having bad generalization quality due to overfitting, datasets are splitted into three sets including, training, validation and test. The CE value of validation set is under the watch during training. Every time the error on the validation set improves, a copy of the model parameters is stored. When the training algorithm terminates, we return these parameters, rather than the latest parameters. In this way, a model with better validation error and hopefully better test error is obtained. Altogether, the algorithm terminates when no parameters have improved over the best recorded validation error for a number of epochs (Also known as patience epochs), or training has reached the maximum allowed number of epochs. 50% of samples were randomly selected for the training set and 25% for validation set and the remaining 25% were for the test set. Exact size of these segments and other properties of these datasets are summarized in Table 1.

Table 1: datasets and their characteristics.

| Dataset | Input features | Output classes | Training instances | Validation instances | Test instances | Total |
|---|---|---|---|---|---|---|
| Diabetes | 8 | 2 | 384 | 192 | 192 | 768 |
| Yeast | 8 | 10 | 742 | 371 | 371 | 1,484 |
| Letter | 16 | 26 | 10,000 | 5,000 | 5,000 | 20,000 |
| Waveform | 21 | 3 | 2,500 | 1,250 | 1,250 | 5,000 |
| Mushrooms | 21 | 2 | 4,062 | 2,031 | 2,031 | 8,124 |

The input data is normalized feature-based, each feature has the range of a mean of zero and unit. Input value for the bias vector is 1 and the network weight connections are initialized by random values in the range [–1, 1] with the same random seed. Also, the learning rate in SGD algorithm is set to 0.1.

### 4.2 The result analysis and discussion

The procedure is performed for Waveform problem from Table 1, and contribution percentage as a measure of saliency for all 21 input features is visualized using a bar chart in Fig. 2. As this figure shows, features 10 and 11 are most salient features while features such as, 1 and 2, have relatively very low contribution in model output and they can be eliminated from the dataset. In order to verify the effectiveness of the proposed measure of saliency, two approaches are assessed:

- First approach: Features firstly are sorted by their saliency in ascending order. By starting from the state of having feature 1 (having lowest contribution) alone in input data, other features are step-wisely attached to the input data, and for each step a new network initialized to be trained with the new constructed dataset. Fig. 3.a demonstrates validating and testing accuracy of the network for each step in Waveform problem.

- Second approach: Features in this approach are conversely sorted in descending order. First by considering feature 11 (as the most salient feature), alone in input data, other features are attached step by step, and for each step a new network is initialized and trained with the new dataset similar to the previous approach. Fig. 3.b demonstrates validating and testing accuracy of the network for each step in Waveform problem.

As Fig. 3. shows, convergence of classification accuracy in the second approach which adds features by descending order is way quicker than the other approach. In fact, generated FNN models that are trained over data compromising the features with most saliency have achieved extremely higher accuracy. This proves the proposed method provides a compelling ranking system for input features. The simulation demonstrates that in case of Waveform problem, the optimized dataset compromising only 11 features is sufficient for the classification task.

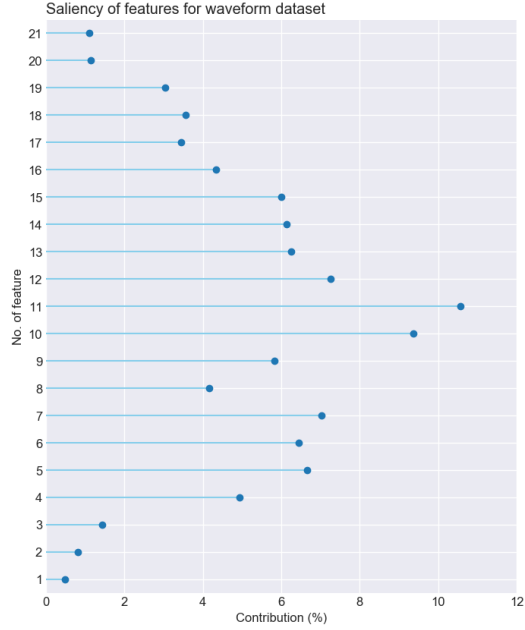

Figure 2: Contribution percentage of input features in Waveform problem.

In the same manner, two approaches are used for the Mushrooms problem. Fig. 4. contains the result which indicates that using as low as 7 features selected from this dataset is enough in order to achieve 100% classification accuracy. In addition, here much alike Waveform experiment, it is verified that the method provides a promising solution for feature selection and specifically detecting irrelevant features.

The procedure is performed for other problems as well, and the final size of the selected subset of features and testing classification accuracy of the network with the optimized dataset are summarized in Table 2. It can be stated from the result that the classification accuracy after significant feature reduction is higher or the same in Yeast, Letter, and Waveform, Mushrooms. Although performance of the FNN over Diabetes dataset is slightly lower than the state of presence of all features, but significant amount (75%) of features are removed, which still indicates good accuracy.

The presented method can be viewed as a wrapper method according to the official definition of wrapper explained in the first section, which emphasizes on employing a learner in the feature selection process.

SA also can be used in the same way for regression problems, and relative importance of features can be revealed with respect to the model with a single output that tries to approximate a desired function. The more a feature has contribution to model output variance, the more it is considered to be important. By normalizing the calculated contribution value assigned to that feature among other features, the percentage value of importance of the feature is relatively determined regarding other features. In fact, for the purpose of the generalization, any function approximate models in machine learning area with continuous or discrete output capable of being studied by SA, could be used for such feature selection method.

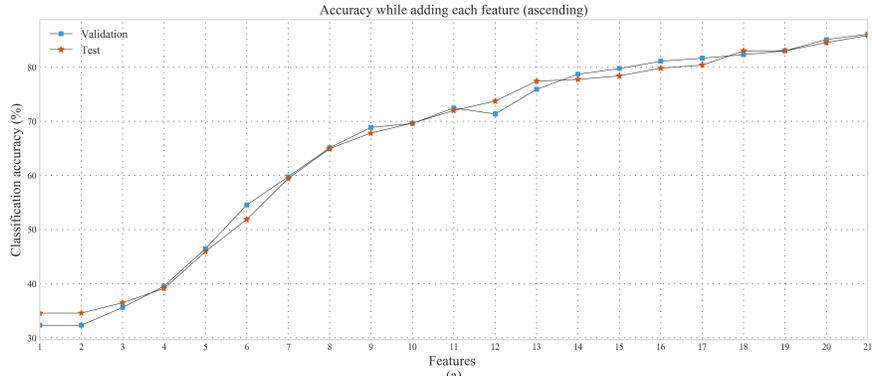
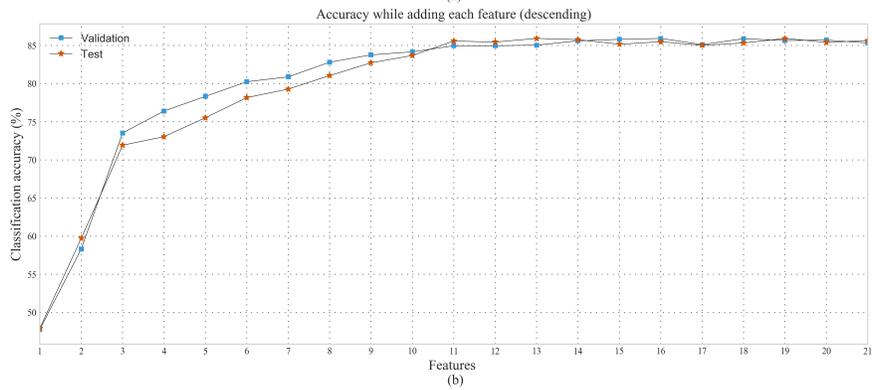

Figure 3: Classification accuracy in each step for Waveform problem.

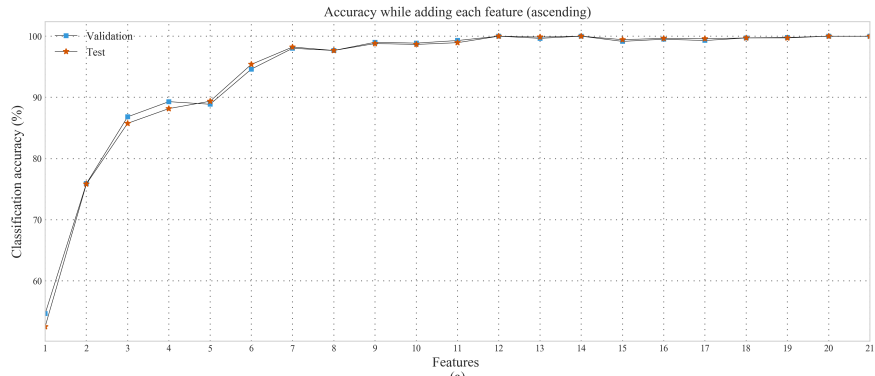
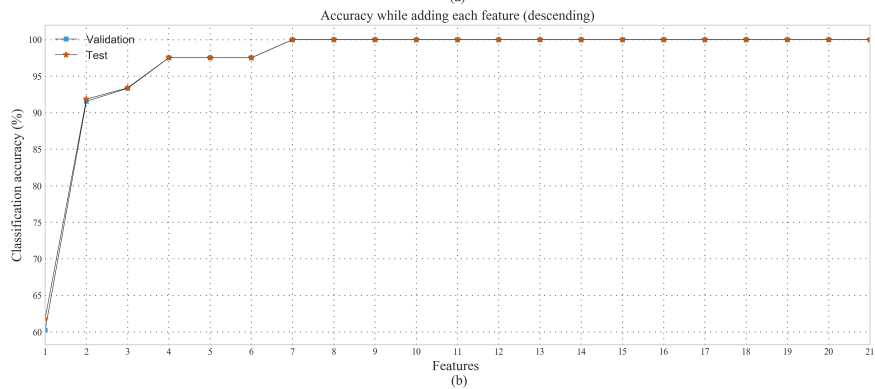

Figure 4: Classification accuracy in each step for Mushroom problem.

Table 1: final size of the selected subset of features and testing classification accuracy of the network with the optimized dataset.

| Dataset | Unprocessed dataset | | Optimized dataset | |
| --- | --- | --- | --- | --- |
| | Features | Accuracy (%) | Features | Accuracy (%) |
| Diabetes | 8 | **78.65** | 2 | 78.12 |
| Yeast | 8 | 59.29 | 6 | **59.56** |
| Letter | 16 | 85.38 | 11 | **85.82** |
| Waveform | 21 | 85.04 | 11 | **85.44** |
| Mushrooms | 21 | **100.00** | 7 | **100.00** |

# 5    Conclusion and future directions

In this work, a new measure of saliency is introduced that provides a valuable ranking system for features using sensitivity analysis and a multi-output feedforward neural network. Sensitivity analysis reveals the influence of features in model output variance. Investigating the relative contribution of features to every single network output unit or in other words, each class label, helps in quantifying the relative importance of each feature concerning the resultant model output. ultimately, this leads to the detection of redundant features in input data. Eliminating these features reduces CPU time as the dimensionality of the input data decreases. As the experimental results also demonstrated, ML model achieved better performance over the optimized datasets.

It should be noted that the presented ranking system for features significantly provides a good assistance for a threshold-based feature selection task, but since the feature selection genuinely is a combinatorial optimization problem, it can be employed along other optimization methods, such as evolutionary computation. This allows to select the best combination of most important features. For instance, using the proposed criteria in heuristic information of a meta-heuristic algorithm can be a good approach in order to omit the procedure of manually selecting the features, and assigning this task completely to the algorithm itself.